\documentclass[letterpaper, 10 pt, conference]{ieeeconf}   
\IEEEoverridecommandlockouts                                
\usepackage{epsfig}  
\usepackage{amsmath}  
\usepackage{url} 
\usepackage{xspace} 
\usepackage{graphicx} 
\usepackage{comment} 
\usepackage{tabu}
\usepackage[table]{xcolor}
\usepackage{caption}
\definecolor{tableShade}{gray}{0.95} 

\usepackage{color} 
\definecolor{darkgreen}{rgb}{0,.75,0}

\title{\LARGE \bf 
Introducing SLAMBench, a performance and accuracy \\benchmarking methodology for SLAM
} 
 
\author{Luigi Nardi$^{1}$, Bruno Bodin$^{2}$, M. Zeeshan Zia$^{1}$, John Mawer$^{3}$, Andy Nisbet$^{3}$, Paul H. J. Kelly$^{1}$, \\
Andrew J. Davison$^{1}$, Mikel Luj\'an$^{3}$, Michael F. P. O'Boyle$^{2}$, Graham Riley$^{3}$, Nigel Topham$^{2}$ and Steve Furber$^{3}$\\ \\ \vspace{2em}
\thanks{$^{1}$Dept. Computing, Imperial College,  London, UK {\tt\small \{luigi.nardi, zeeshan.zia, p.kelly, ajd\}@imperial.ac.uk}} 
\thanks{$^{2}$Institute for Computing Systems Architecture, The Univ. of Edinburgh, Scotland 
        {\tt\small \{bbodin, mob, npt\}@inf.ed.ac.uk}} 
\thanks{$^{3}$School of Computer Science, Univ. of Manchester, UK 
        {\tt\small \{andy.nisbet, john.mawer, mikel.lujan, graham.riley, steve.furber\}@manchester.ac.uk}} 
} 
\begin{document} 
\makeatletter
\let\@oldmaketitle\@maketitle
\renewcommand{\@maketitle}{\@oldmaketitle
\vspace{-3em}
  \includegraphics[width=\linewidth,height=7\baselineskip]
    {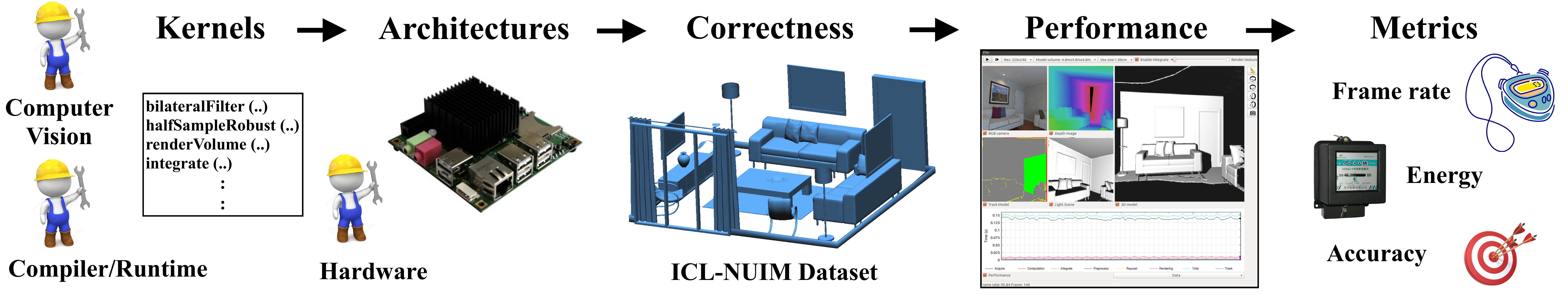}
\captionof{figure}{
\label{figTeaser} 
The SLAMBench framework makes it possible for experts
coming from the computer vision, compiler, run-time, and hardware communities 
to cooperate in a unified way to tackle algorithmic and implementation alternatives.}\bigskip}
\makeatother
\maketitle 
\thispagestyle{empty} 
\pagestyle{empty} 

\setcounter{figure}{1}

\begin{abstract} 
Real-time dense computer vision and SLAM offer great potential for a new level of scene modelling, 
tracking and real environmental interaction for many types of robot, but their high computational requirements mean that use on mass market embedded platforms is challenging. 
Meanwhile, trends in low-cost, low-power processing are towards massive parallelism and heterogeneity, 
making it difficult for robotics and vision researchers to implement their algorithms in a performance-portable way.
In this paper we introduce SLAMBench, a publicly-available software framework which represents a starting point for quantitative, comparable and validatable experimental research  
to investigate trade-offs in performance, accuracy and energy consumption of a dense RGB-D SLAM system.  
SLAMBench provides a KinectFusion implementation in 
C++, OpenMP, OpenCL and CUDA, and harnesses the ICL-NUIM dataset of synthetic RGB-D sequences with trajectory and scene ground truth for reliable accuracy comparison of different implementation and algorithms.
We present an analysis and breakdown of the constituent algorithmic elements of KinectFusion, 
and experimentally investigate their execution time on a variety of multicore and GPU-accelerated platforms.  
For a popular embedded platform, we also present an analysis of energy efficiency for different configuration alternatives.
\end{abstract} 
 
\section{INTRODUCTION} 
Recent advances in computer vision have led to real-time algorithms for  
dense 3D scene understanding, such as KinectFusion~\cite{2011Newcombe}, which  
estimate the pose of a depth camera while
building a highly detailed 3D  
model of the environment. Such real-time 3D scene understanding capabilities can radically 
change the way robots interact with and manipulate the world.

While classical point feature-based \textit{simultaneous localisation and mapping} (SLAM) techniques are
now crossing into mainstream products via embedded implementation in projects like Project Tango \cite{ProjectTango} 
and Dyson 360 Eye \cite{DysonLab}, dense SLAM algorithms with their high
computational requirements are largely at the prototype stage on
PC or laptop platforms.
Meanwhile, there has been a great focus in computer vision on
developing benchmarks for accuracy comparison,
but not on analysing and characterising the 
envelopes for performance and energy consumption. 

The growing interest in real-time applications 
means that the difficulty of parallelisation  
affects every application developer. 
Existing benchmarks are typically general purpose,  
e.g. the MediaBench suite \cite{1997Mediabench}  
for evaluation of embedded multimedia and communications, the DaCapo benchmark \cite{2006Dacapo} 
for sequential client and server Java code, and the SPEC CPU benchmark \cite{2006Spec} which stresses the processor, 
memory subsystem and compiler.
Such benchmarks have proved valuable for the evaluation of existing scientific and engineering applications,  
but they fail to capture essential elements of modern computer vision algorithms and,   
in particular, SLAM applications. To address these issues, we present SLAMBench \cite{SLAMBench} (see Fig. \ref{figTeaser}) 
which is a portable implementation of KinectFusion with verifiable accuracy relative to a known ground truth. 
SLAMBench allows computer vision researchers to provide and evaluate alternative algorithmic implementations but 
it also enables domain-specific optimisation, auto-tuning and domain-specific languages (DSLs) in dense SLAM. 
It takes a step forward improving the way academia and industry evaluate  
system design and implementation in computer vision.  
To the best of our knowledge, SLAMBench is the first performance, energy and accuracy benchmark dedicated to 3D scene understanding applications.  

The contributions of this paper are: 
\begin{itemize} 
\item We present a SLAM performance benchmark that combines a framework 
for quantifying quality-of-result with instrumentation of execution 
time and energy consumption.
\item We present a qualitative analysis of the parallel patterns underlying 
each of the constituent computational elements of the KinectFusion algorithm. 
\item We present a characterisation of the performance of the 
KinectFusion implementation using both multicore CPUs and GPUs, 
on a range of hardware platforms from desktop to embedded.
\item We present an analysis of the algorithm's energy consumption under various 
implementation configurations using an embedded, multicore system on chip (SoC) with GPU acceleration. 
\item We offer a platform for a broad spectrum of future research 
in jointly exploring the design space of algorithmic and implementation-level optimisations.
\end{itemize} 
 
\section{RELATED WORK} 
\label{RELATED} 
Computer vision research has traditionally focused on optimising the accuracy of algorithms. 
In autonomous driving, for example, the KITTI benchmark suite \cite{2012Geiger} provides data and  
evaluation criteria for the stereo, optical flow, visual odometry and 3D object recognition.  
The ICL-NUIM dataset \cite{2014Handa} and TUM RGB-D benchmark \cite{2012Sturm} aim to benchmark 
the accuracy of visual odometry and SLAM algorithms.  
However, in an energy and performance constrained context, such as a battery-powered robot,
it is important to achieve sufficient accuracy 
while maximising battery life. 
New benchmarks are needed which provide tools and techniques to investigate these constraints.  
An important early benchmark suite for performance evaluation entirely dedicated to computer vision is SD-VBS \cite{2009Sdvbs}.  
SD-VBS provides single-threaded C and  
MATLAB implementations of 28 commonly used   
computer vision kernels that are combined to build 9 high-level vision applications; 
only some modules relevant to SLAM are included, notably Monte Carlo localisation.
SD-VBS \cite{2009Sdvbs} prohibits modifications to the algorithm, only allowing the implementation to be tuned to suit novel hardware architectures.
This limits the use of the benchmark in the development of novel algorithms. 

Another attempt at such performance evaluation is MEVBench \cite{2011Mevbench}, 
which focuses on a set of visual recognition applications including face detection,  
feature classification, object tracking and feature extraction. 
It provides single and multithreaded C++ implementations for some of the kernels with a special emphasis on low-power embedded systems.  
However, MEVBench only focuses on recognition algorithms and does not include a SLAM pipeline. 

While such efforts are a step in the right direction, 
they do not provide the software tools for accuracy verification and 
exploitation of hardware accelerators or {\em graphics processor units} (GPUs). Nor do they enable investigation of energy consumption, 
performance and accuracy envelopes for 3D scene reconstruction algorithms 
across a range of hardware targets.  
The lack of benchmarks 
stems from the difficulty in systematically comparing the accuracy of the reconstruction while measuring the performance.  
In this work we focus specifically on SLAM and introduce a publicly-available framework for quantitative, 
comparable and validatable experimental research in the form of a benchmark for dense 3D scene understanding.  
A key feature of SLAMBench is that it is designed on top of the recently-proposed ICL-NUIM accuracy benchmark \cite{2014Handa}, and  
thus supports wider research in hardware and software. 
The quantitative evaluation of solution accuracy into SLAMBench enables algorithmic research to be performed.
The latter is an important feature that is lacking in current performance benchmarks.  
A typical output of SLAMBench consists of the performance achieved, and the accuracy of the result 
along with the energy consumption (on platforms where such measurement is possible).
These parameters capture the potential trade-offs for real-time vision  platforms.  
 
\section{SLAM AND KINECTFUSION} 
\label{SLAM} 
SLAM systems aim to perform real-time localisation and mapping 
``simultaneously'' for a sensor moving through an unknown environment. 
SLAM could be part of a mobile robot, enabling the robot to update its estimated position in an environment,  
or an augmented reality (AR) system where a camera is moving through the environment, 
allowing the system to render graphics at appropriate locations in the scene 
(e.g. to illustrate repair procedures in-situ \cite{2011Henderson} or 
to animate a walking cartoon character on top of a table).
Localisation typically estimates the location and pose of the sensor (e.g. a camera) with regard to a map  
which is extended as the sensor explores the environment. 
 
The KinectFusion \cite{2011Newcombe} algorithm  utilises a  
depth camera to perform real-time localisation and dense mapping.   
A single raw depth frame from these devices has both noise and holes. 
KinectFusion registers and fuses the stream of measured depth frame obtained as the scene is viewed from different viewpoints into a clean 3D geometric map. 
While it is beyond the scope of this paper to go into the details of the KinectFusion algorithm,  
we briefly outline the key computational steps involved in the top row of Fig.~\ref{figFramework}.  
KinectFusion normalizes each incoming depth frame and  applies a bilateral filter \cite{1998Tomasi} (\textit{Preprocess});  
before computing a point cloud (with normals) for each pixel in the camera frame of reference.  
Next, KinectFusion estimates (Track) the new 3D pose of the moving camera by registering 
this point cloud with the current global map using a variant of \textit{iterative closest point} (ICP) \cite{1992Besl}. 
Once the new camera pose has been estimated, the corresponding depth map 
is fused into the current 3D reconstruction (\textit{Integrate}).  
KinectFusion utilises a voxel grid as the data structure to represent the map,  
employing a \textit{truncated signed distance function} (TSDF) \cite{1996Curless} to represent 3D surfaces.  
The 3D surfaces are present at the zero crossings of the TSDF and can be recovered by a raycasting step,  
which is also useful for visualising the reconstruction.  
The key advantage of the TSDF representation is that it 
simplifies fusion of new data into existing data to the calculation of a 
running average over the current TSDF volume and the new depth image. 
 
KinectFusion has been adopted as a major building block in more recent SLAM systems \cite{2012Whelan,2013Keller,2013Zeng,2013Chen,2013Niessner} and 
is part of Microsoft's Kinect SDK \cite{2014MicrosoftKinectSDK}.
A number of open implementations of the KinectFusion algorithm have been made in recent years, 
including KFusion \cite{KFusion} (utilised in this paper) and KinFu \cite{2011PCL}.  

\section{SLAMBENCH} 
\label{SLAMBENCH} 
\begin{figure}[t] 
\includegraphics[width=1.0\columnwidth]{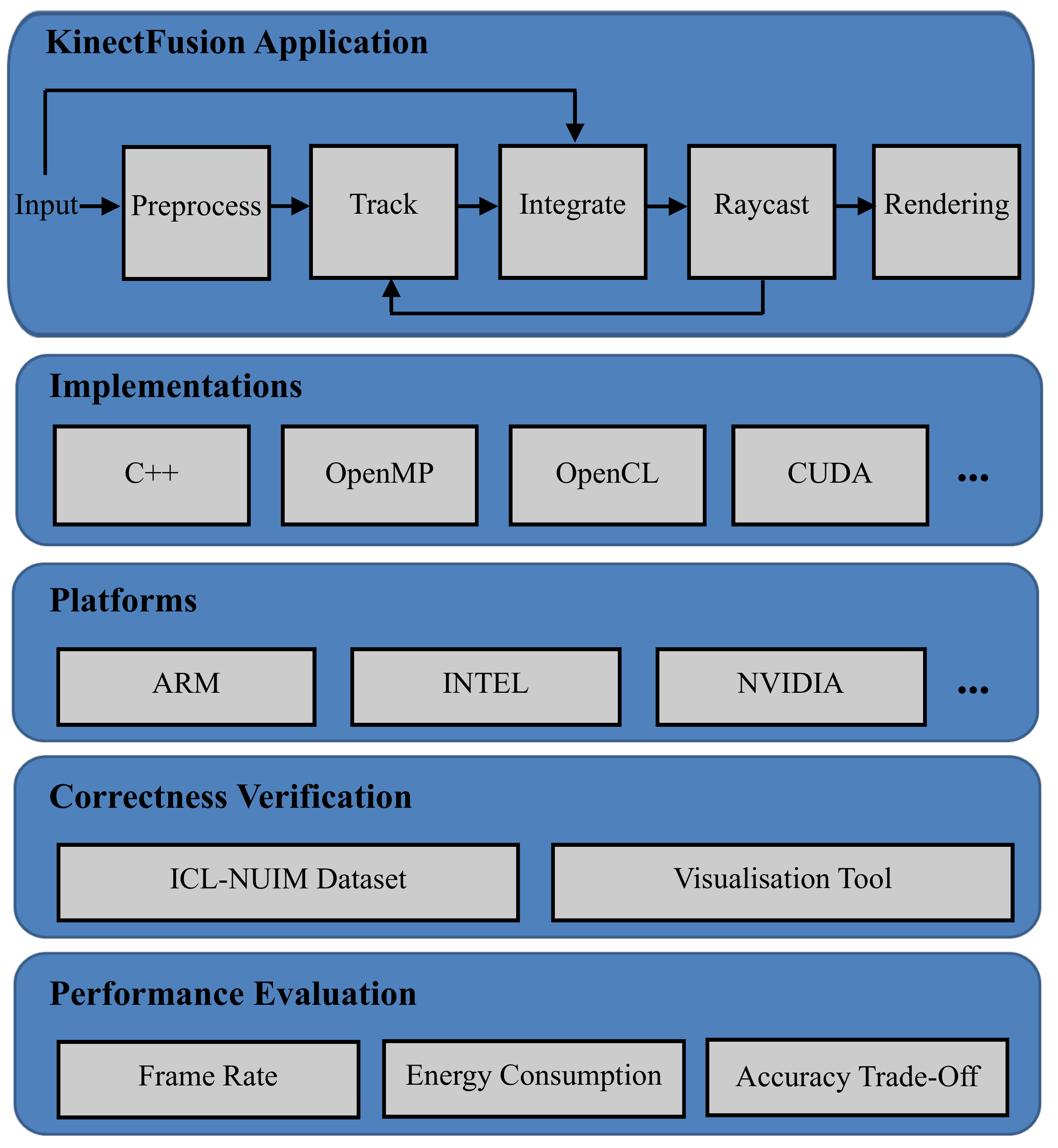} 
\caption{SLAMBench framework.}
\label{figFramework} 
\end{figure}

SLAMBench, see Fig.~\ref{figFramework}, is a benchmark that provides portable, but untuned, KinectFusion \cite{2011Newcombe} implementations in C++ (sequential),  OpenMP, CUDA and OpenCL for a range of target platforms.   
As an example, the ODROID board contains two GPU devices, see Sec.~\ref{secDevices}, but the current OpenCL implementation only exploits one GPU device.  SLAMBench includes techniques and tools to validate an implementation's accuracy using the   ICL-NUIM dataset along with algorithmic modifications to explore accuracy and performance trade-offs.  
 
\subsection{ICL-NUIM DATASET} 
\label{ICL-NUIM} 
ICL-NUIM \cite{2014Handa} is a high-quality synthetic dataset providing RGB-D sequences for 4 different camera trajectories through a living room model. 
An {\em absolute trajectory error} (ATE) is calculated 
as the difference between the ground truth and the estimated trajectory of a camera produced by a SLAM implementation, 
this enables system accuracy to be measured at each frame. 

Fig.~\ref{figTeaser} shows the input scene from the ICL-NUIM dataset and the reconstruction performed using SLAMBench.  
ICL-NUIM can provide frames sequences that: (1), are free of noise, and (2), 
have noise added to RGB-D frames according to a statistical model matching Microsoft's Kinect sensor.
The noise model is applied to each frame used as input to a SLAM system in order to give
realistic results.  
 
The ICL-NUIM benchmark provides not only a number of realistic pre-rendered sequences,  
but also open source code that can be used by researchers in order to generate their own test data as required.  
As the unmodified ICL-NUIM dataset is the input of the SLAMBench benchmark any user can run SLAMBench in a straightforward way on other datasets.   

\subsection{SLAMBENCH KERNELS} 
\label{SLAMBENCHKERNELS} 
\begin{table}[bt]
\begin{center}
\begin{tabu}{cccccc}
\hline \noalign{\smallskip}
\rowfont{\bfseries\footnotesize} Kernels & Pipeline & Pattern & In & Out & \%\\
\noalign{\smallskip} \hline
\hline 
\rowfont{\footnotesize}acquire & Acquire & n/a & pointer & 2D & 0.03\\
\rowfont{\footnotesize}mm2meters & Preprocess & Gather & 2D & 2D & 0.06\\
\rowfont{\footnotesize}bilateralFilter & Preprocess & Stencil & 2D & 2D & 33.68\\
\rowfont{\footnotesize}halfSample & Track & Stencil & 2D & 2D & 0.05\\
\rowfont{\footnotesize}depth2vertex & Track & Map & 2D & 2D & 0.11\\
\rowfont{\footnotesize}vertex2normal & Track & Stencil & 2D& 2D & 0.27\\
\rowfont{\footnotesize}track & Track & Map/Gather & 2D & 2D & 4.72\\
\rowfont{\footnotesize}reduce & Track & Reduction & 2D & $6$x$6$ & 2.99 \\
\rowfont{\footnotesize}solve & Track & Sequential & $6$x$6$ & $6$x$1$ & 0.02\\
\rowfont{\footnotesize}integrate & Integrate & Map/Gather & 2D/3D & 3D & 12.85 \\
\rowfont{\footnotesize}raycast & Raycast & Search/Stencil & 2D/3D & 2D & 35.87 \\
\rowfont{\footnotesize}renderDepth & Rendering & Map & 2D & 2D & 0.12 \\
\rowfont{\footnotesize}renderTrack & Rendering & Map & 2D & 2D & 0.06\\
\rowfont{\footnotesize}renderVolume & Rendering & Search/Stencil & 3D & 2D & 9.18\\
\hline
\end{tabu}
\end{center}
\caption{SLAMBench kernels. 
\textit{Pipeline} refers to SLAM high-level building blocks of the top row of Fig~\ref{figFramework}.
\textit{Pattern} describes the parallel computation pattern (see sec.~\ref{secParallelPatterns}) of a kernel. 
\textit{\%} gives the percentage of the time spent in the kernel while running sequentially on the Exynos 5250 processor (see platforms Table~\ref{tabPlatforms}). 
For the breakdown in other platforms see Fig.~\ref{figKernel}.
}
\label{tabKernels} 
\end{table}
Table \ref{tabKernels} summarises the 14 main computationally significant computer vision kernels contained in SLAMBench.
The following is a description of each kernel:
\begin{enumerate}
\item \textbf{acquire}: acquires a new RGB-D frame. 
This input step is included explicitly in order to account for I/O costs
 during benchmarking, and for real applications.
\item \textbf{mm2meters}: transforms a 2D depth image from millimeters to meters. 
If the input image size is not the standard $640$x$480$, only a part of the image is converted and mapped into the output. 
\item \textbf{bilateralFilter}: is an edge-preserving blurring filter applied to the depth image. 
It reduces the effects of noise and invalid depth values.
\item \textbf{halfSample}: is used to create a three-level image pyramid by sub-sampling the filtered depth image. Tracking solutions from low resolution images in the pyramid are used as guesses to higher resolutions. 
\item \textbf{depth2vertex}: transforms each pixel of a new depth image into a 3D point (vertex). 
As a result, this kernel generates a point cloud.
\item \textbf{vertex2normal}: computes the normal vectors for each vertex of a point cloud. 
Normals are used in the projective data association step of the ICP algorithm to calculate the point-plane distances 
between two corresponding vertices of the synthetic point cloud and a new point cloud. 
\item \textbf{track}: establishes correspondence between vertices in the synthetic and new point cloud.
\item \textbf{reduce}: adds up all the distances (errors) of corresponding vertices of two point clouds for the minimisation process.
On GPUs, the final sum is obtained using a parallel tree-based reduction. 
\item \textbf{solve}: uses TooN \cite{TooN} to perform a singular value decomposition on the CPU that solves a $6$x$6$ linear system. A  6-dimensional vector is produced to correct the new estimate of camera pose. 
\item \textbf{integrate}: integrates the new point cloud into the 3D volume. 
It computes the running average used in the fusion described in Sec.~\ref{SLAM}. 
\item \textbf{raycast}: computes the point cloud and normals corresponding to the current estimate of the camera position. 
\item \textbf{renderDepth}: visualises the depth map acquired from the sensor using a colour coding.
\item \textbf{renderTrack}: visualises the result of the tracking. 
For each pixel different colours are associated with one of the possible outcomes of the tracking pass, 
e.g. `correct tracking', `pixel too far away', `wrong normal', etc.
\item \textbf{renderVolume}: visualises the 3D reconstruction from a fixed viewpoint (or a user specified viewpoint when in the GUI mode). 
\end{enumerate}
All kernels are implemented on GPUs, in OpenCL and CUDA, except \textit{acquire} and \textit{solve}.

In addition to these kernels, the SLAMBench pipeline also contains two initialisation kernels 
not shown in Table~\ref{tabKernels} and Fig.~\ref{figFramework}: \textit{generateGaussian} and \textit{InitVolume}. 
These kernels are part of an initialisation step that is performed only at start-up. 
\textit{generateGaussian} generates a Gaussian bell curve and stores it in a 1D array; 
\textit{initVolume} initialises the 3D volume. 

\subsection{SLAMBENCH PARALLEL PATTERNS}
\label{secParallelPatterns}
\begin{figure*}[tb]
\parbox{0.40\columnwidth}{\center \small (a)} \hfill 
\parbox{0.40\columnwidth}{\center \small (b)}
\parbox{0.40\columnwidth}{\center \small (c)}
\parbox{0.40\columnwidth}{\center \small (d)}
\parbox{0.40\columnwidth}{\center \small (e)}\\
\includegraphics[width=0.40\columnwidth]{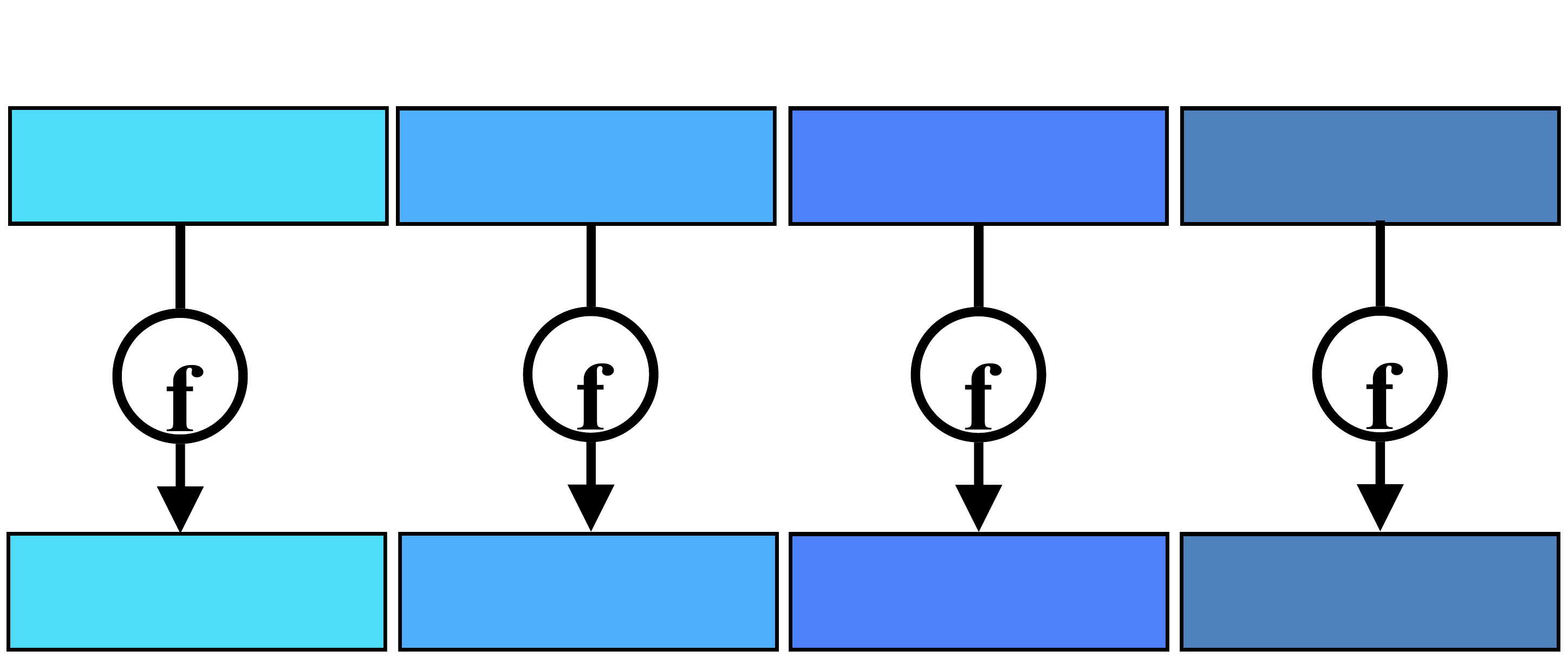}
\includegraphics[width=0.40\columnwidth]{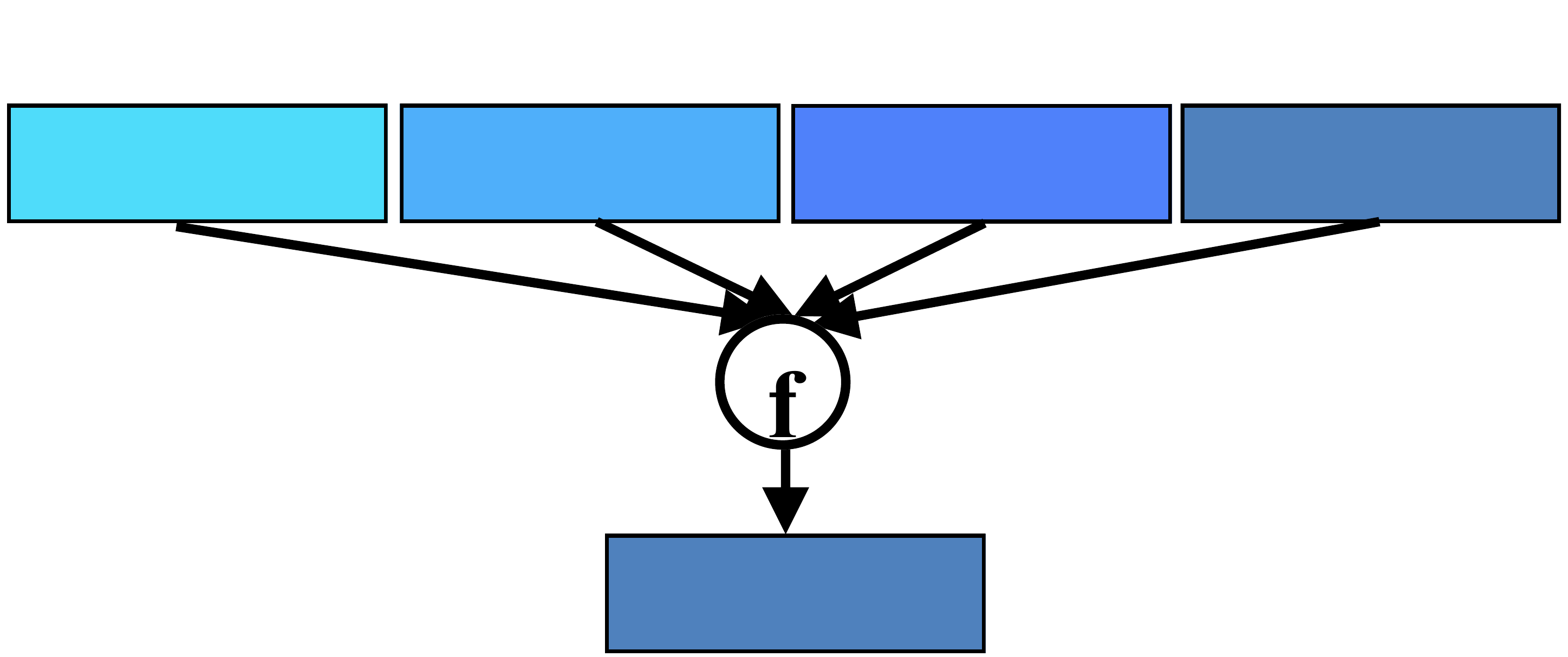}
\includegraphics[width=0.40\columnwidth]{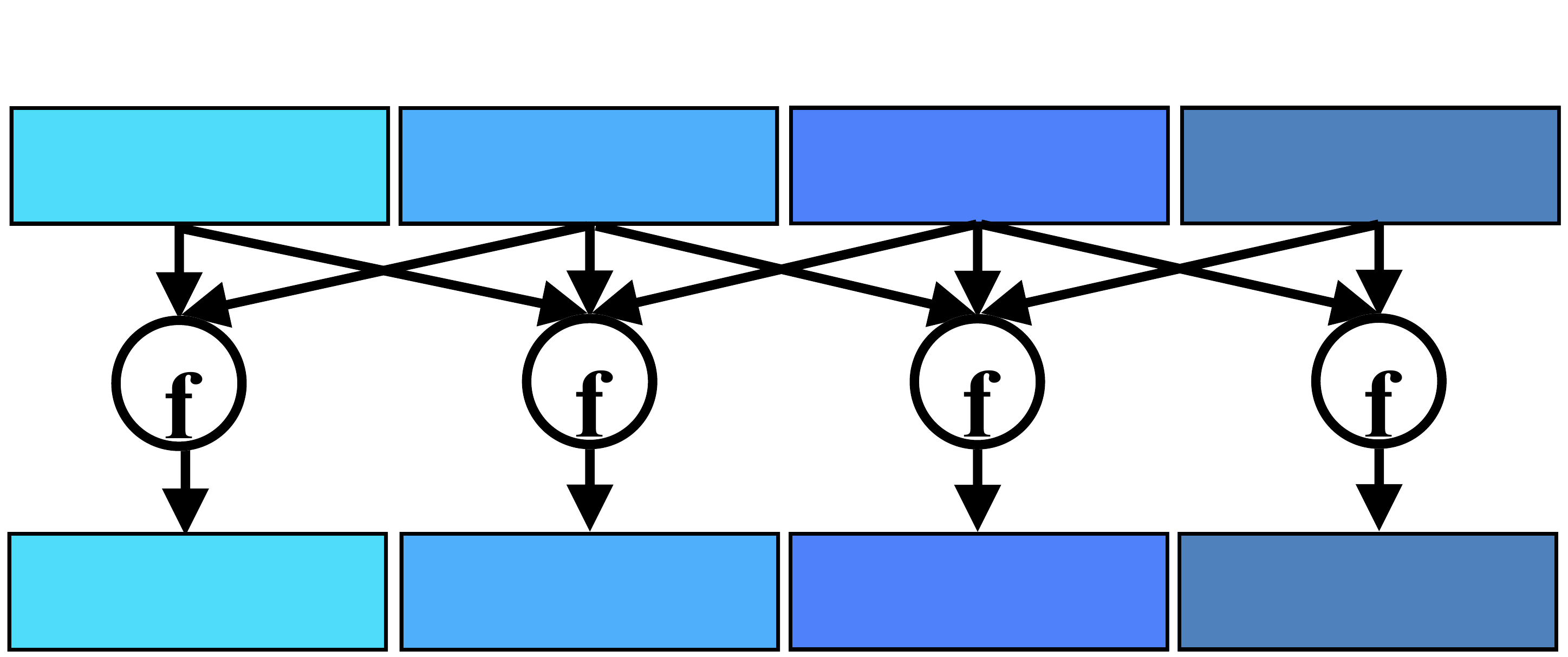}
\includegraphics[width=0.40\columnwidth]{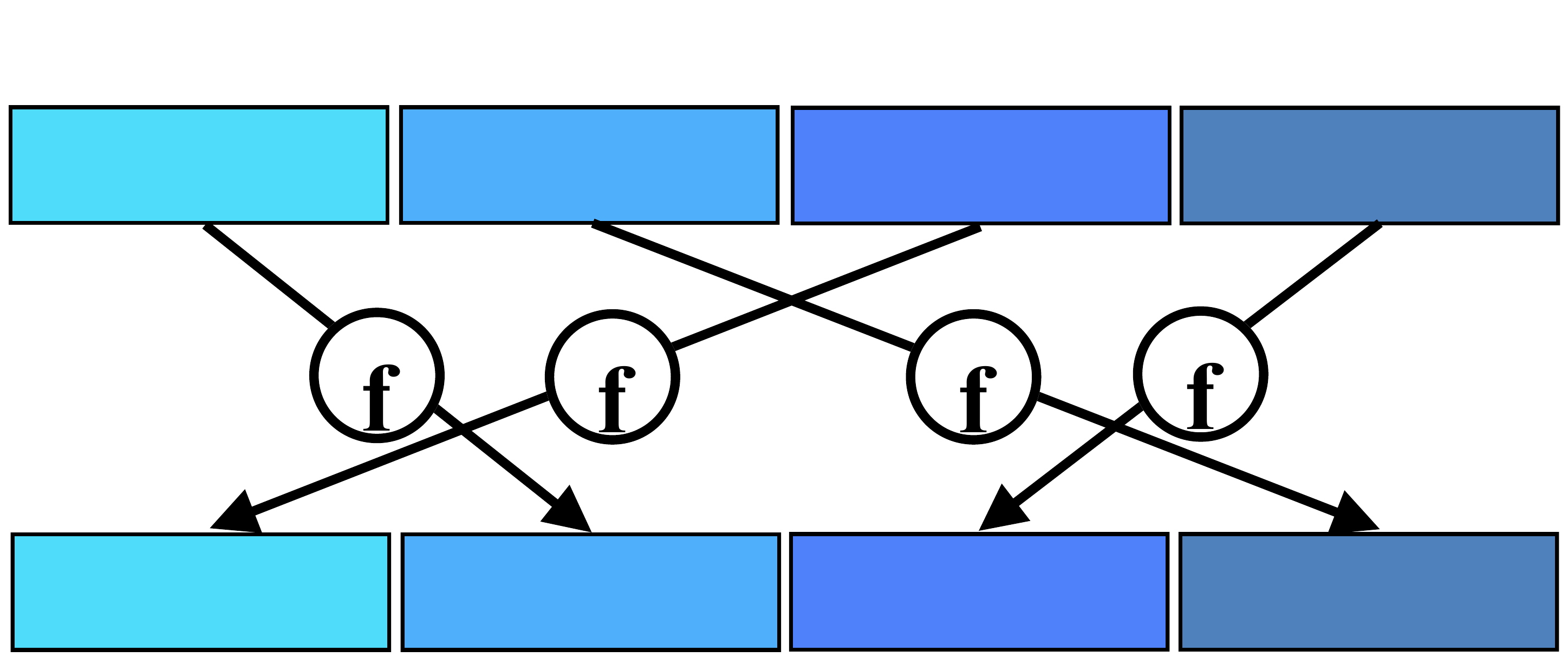}
\includegraphics[width=0.40\columnwidth]{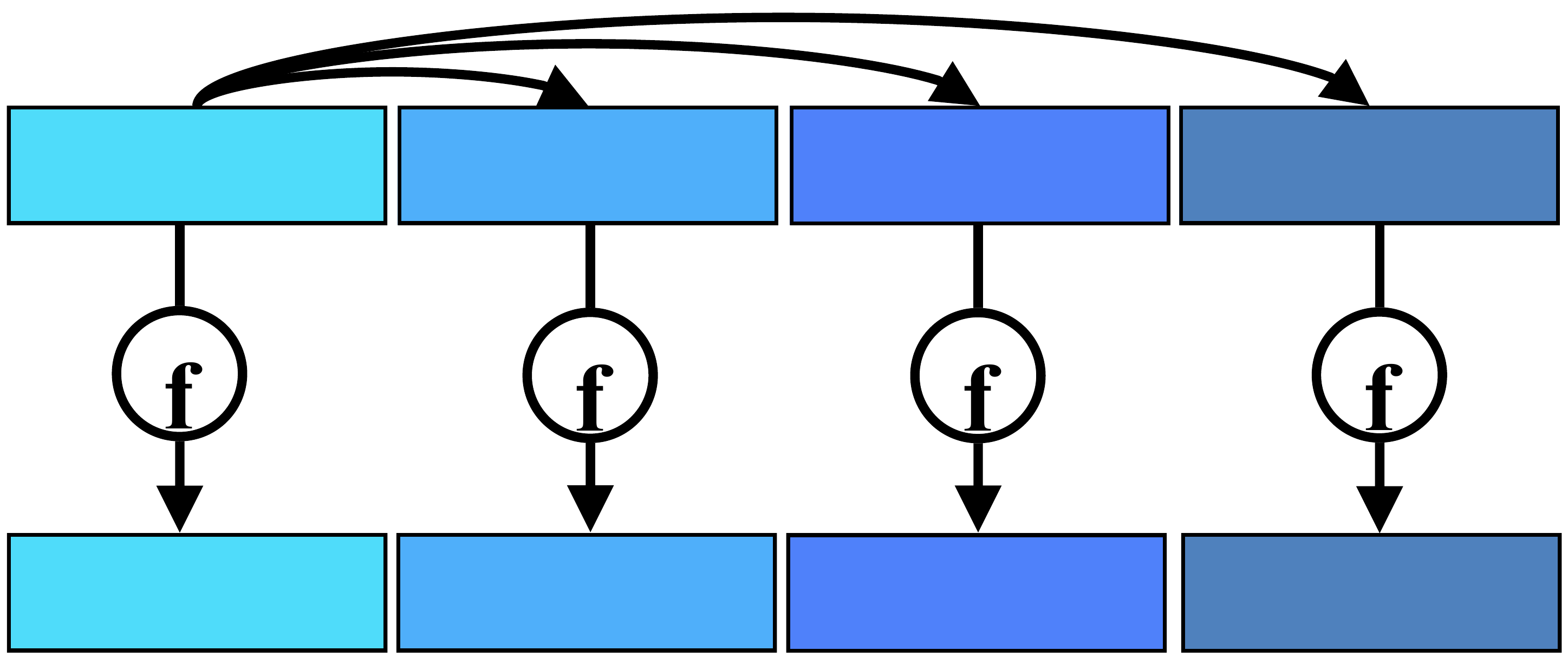}
\caption{Parallel patterns: (a) Map, (b) Reduction, (c) Stencil, (d) Gather and (e) Search.}
\label{figParallelPatterns}
\end{figure*}
The parallel patterns of Table~\ref{tabKernels} are defined following definitions from \cite{2010Mccool} and   
shown in Fig.~\ref{figParallelPatterns}. 
Since the kernels represent complex real-world SLAM computations, the computation of a single kernel may fit multiple parallel patterns as depicted in Table~\ref{tabKernels}. 
\begin{itemize}
\item \textbf{Map}: (see Fig.~\ref{figParallelPatterns}(a)),  a pure function operates on every
element of an input array and produces one result per element.
Threads do not need to synchronise.
\item \textbf{Reduction}: (see Fig.~\ref{figParallelPatterns}(b)), a reduction function combines all the elements
of an input array to generate a single output. 
In this case, tree-based implementations can be used to parallelise such a reduction. 
\item \textbf{Stencil}: (see Fig.~\ref{figParallelPatterns}(c)), each output element is computed
by applying a function on its corresponding input
array element and its neighbors.
\item \textbf{Gather}: (see Fig.~\ref{figParallelPatterns}(d)), is similar to map but its memory accesses are random. 
Parallel gather implementations are similar to those of map. 
\item \textbf{Search}: (see Fig.~\ref{figParallelPatterns}(e)), is
similar to gather, but it retrieves data from an array based on matching against content.
Parallelism is achieved by searching for all keys in parallel, or by searching in different parts of the array in parallel.
\end{itemize}
\subsection{PERFORMANCE EVALUATION METHODOLOGY} 
\label{PERFORMANCE} 
A methodology for SLAM performance evaluation is necessary for three reasons.  
First, since we are dealing with numerical approximations and 
iterative algorithms a systematic way to  check the accuracy of the results on different platforms is required. 
For example, in the pose estimation, KFusion assumes small motion, 
through a small angle approximation,  
from one frame to the next in order to achieve fast projective data association \cite{1992Besl}.
Also, the rigid body transform in the camera pose estimation is computed  
by testing whether each new iteration carries new useful information.  
Finally, the error between point clouds is computed using a reduction of $l^{2}-norms$, 
whose result may depend on the order in which summation takes place.
 
The second issue requiring careful evaluation methodology
is that we are dealing with an application that may be used in different contexts, such as mobile robotics and AR.  
Each application places different stress on the SLAM pipeline.  
 
Thirdly, the KinectFusion computation is dependent on the images acquired, the way the camera is moved and the tracking frame rate.  
A methodology is required to enable a SLAMBench user to avoid divergent behaviors 
across different platforms where the target system cannot process all frames in real-time. 
We use a dataset of pre-recorded scenes and the 
\textit{Process-every-frame} mode described below. 
 
The SLAMBench methodology consists of four elements: 
\subsubsection{GUI vs terminal interface} 
\label{GUI} 
SLAMBench includes raycasting for visualisation but we have not included the display of the image.  
Rendering using OpenGL lies outside the scope of this work.
\subsubsection{Pre-recorded scenes} 
\label{Pre-recorded} The KinectFusion computation depends upon the scene and camera trajectory. 
In order to address the variability that comes with the images and the camera movement in the benchmark space we use pre-recorded scenes described in Sec.~\ref{ICL-NUIM}, thus ensuring reproducibility. 

It is important to notice that using pre-recorded scenes can impact the performance of the application
since in this case the \textit{acquire} is negligible compared to other kernels, as shown in Sec.~\ref{secPerformance}.
But in a real-life context the input device induces unavoidable latency, 
which makes it important to take this into account in the performance analysis.

\subsubsection{Frame rate} 
\label{Framerate} 
in a real-time application context, 
the frame rate of the application depends on the computational speed of the platform used. 
If the application is unable to process frames in real-time then input frames are dropped. 
To ensure consistent evaluation of accuracy, energy consumption and performance 
it is necessary to guarantee that the compared applications process the same set of frames.  
For this reason we introduce a \textit{Process-every-frame} mode, where the application acquires the next pre-recorded frame (without considering timestamps) from a dataset file only when the previous frame has been processed.  
The frames are enumerated and scheduled in same order on all platforms.  
This ensures that all simulations process exactly the same frames  
thus rendering the computation independent of the frequency of the camera and performance of the platform.  
 
\subsubsection{Accuracy evaluation}  
\label{Accuracy}  
SLAMBench provides an automatic tool that performs testing against the ground truth from ICL-NUIM.  
The ATE, mean, median, standard deviation, min and  
max statistics errors (as described in \ref{ICL-NUIM}) are computed using the synthetic trajectories 
given by the ICL-NUIM. 
This allows us to ensure that versions of the implementation using different languages and platforms are sufficiently accurate, whilst not expecting bit-wise output equivalence.
 
We note SLAMBench currently only exploits the trajectory ground truth from
ICL-NUIM, and not the 3D reconstruction it also offers.
This is because, as described  
by Handa et al. \cite{2014Handa}, interactively driven computer vision tools are required in order to perform reconstruction comparison. 
However, localisation and mapping are strongly correlated, which 
means that the ATE serves as a good indication of whether the reconstruction is accurate enough. 
 
\section{Experimental evaluation} 
\label{secEXPERIMENTAL}

\subsection{DEVICES}
\label{secDevices}
For our experiments we used five platforms, Table \ref{tabPlatforms} summarises their characteristics.
\begin{table*}[htb]
\begin{center}
\begin{tabu}{cccccc}
\hline \noalign{\smallskip}
\rowfont{\bfseries\small} Machine names & TITAN & GTX870M & TK1 & ODROID (XU3) & Arndale \\
\noalign{\smallskip} \hline
\hline 
\rowcolor{tableShade}
\textbf{Machine type} & Desktop & Laptop & Embedded & Embedded & Embedded\\
\textbf{CPU} & i7 Haswell & i7 Haswell & NVIDIA 4-Plus-1 & Exynos 5422 & Exynos 5250 \\
\rowcolor{tableShade}
\textbf{CPU cores} & 4 & 4 & 4 (Cortex-A15) + 1 & 4 (Cortex-A15) + 4 (Cortex-A7)& 2 (Cortex-A15) \\
\textbf{CPU GHz} & 3.5 & 2.4 & 2.3 & 1.8 & 1.7 \\
\rowcolor{tableShade}
\textbf{GPU} & NVIDIA TITAN & NVIDIA GTX 870M & NVIDIA Tegra K1 & ARM Mali-T628-MP6 & ARM Mali-T604-MP4\\
\textbf{GPU architecture} & Kepler & Kepler & Kepler & Midgard 2nd gen.& Midgard 1st gen. \\
\rowcolor{tableShade}
\textbf{GPU FPU32s} & 2688 & 1344 & 192 & 60 & 40   \\        
\textbf{GPU MHz} & 837 & 941 & 852 & 600& 533 \\
\rowcolor{tableShade}
\textbf{GPU GFLOPS (SP)} & 4500 & 2520 & 330 & 60+30 (72+36) & 60 (71) \\
\textbf{Language} & CUDA/OpenCL/C++ & CUDA/OpenCL/C++ & CUDA/C++ & OpenCL/C++ & OpenCL/C++\\    
\rowcolor{tableShade}
\textbf{OpenCL version} & 1.1 & 1.1 & n/a & 1.1 & 1.1 \\    
\textbf{Toolkit version} & CUDA 5.5 & CUDA 5.5 & CUDA 6.0 & Mali SDK1.1. & Mali SDK1.1 \\    
\rowcolor{tableShade}
\textbf{Ubuntu OS (kernel)} & 13.04 (3.8.0) & 14.04 (3.13.0) & 14.04 (3.10.24) & 14.04 (3.10.53) & 12.04 (3.11.0) \\    
\hline
\end{tabu}
\end{center}
\caption{Devices used for our experiments.}
\label{tabPlatforms} 
\end{table*}
For the sake of brevity we refer to these machines as TITAN, GTX870M, TK1, ODROID and Arndale. 
SLAMBench is targeted at processors from the desktop down to low power arenas, e.g. processors used in smart-phones and tablets.  
These embedded devices are gaining in popularity and acquiring more capable cameras and mobile processors. 
We carefully select the platforms to be representative of the actual devices used by consumers and industry 
in the different desktop, mobile and embedded categories.

Comparison between different GPU architectures is a non trivial task;  
this is not helped by the wide range of vendor specific languages used, 
commercial sensitivity and varying architectures. 
NVIDIA characterise their devices by the number of ``CUDA cores''. 
Thus the TK1 has 
192 CUDA cores while ARM's GPUs have 4-8 more powerful cores.

In the ARM embedded devices space we run experiments on the Hardkernel ODROID-XU3 based on the Samsung Exynos 5422.  
The Exynos 5422 implements ARM's big.LITTLE heterogeneous multiprocessing (HMP) solution with four Cortex-A15, 
``big'' out of order processors, four ``LITTLE'' in order processors and a Mali-T628-MP6 GPU; 
this GPU essentially consists of two separate devices one consisting of 4 cores and the other 2.  
In our experiments we only use the 4-core GPU device. 
The ODROID platform has integrated power monitors which will be explored in Sec~\ref{secPower}. 
At the ARM end we also perform experiments on the dual-core Arndale board based on a Samsung Exynos 5250 and MALI-T604-MP4 GPU. 

Jetson TK1 is NVIDIA's embedded Linux development platform featuring a Tegra K1 SoC. 
It is provided with a quad-core ARM Cortex-A15 (NVIDIA 4-Plus-1) in the same SoC with an NVIDA GPU 
 composed of 192 CUDA cores. 

At the desktop end of the spectrum we selected a 4-core Intel Haswell i7-4770K 
associated with an NVIDIA GTX TITAN GPU containing 2688 CUDA cores.
This machine is a high-end desktop with a power envelope up to 400 W, 
although the actual power consumption is, in our experience, considerably less. 
Including such a powerful machine in the testbed is interesting in order to explore the limits of the application when many parallel 
compute units are available. 

In order to be as representative as possible on the cross platform evaluation, we also include a high-end laptop. 
The laptop features an Intel Haswell mobile processor i7-4700MQ together with an NVIDIA GTX 870M mobile discrete GPU. 
This GPU contains 1344 CUDA cores,
 making it more suitable than the GTX TITAN for high-end laptop usage.

\subsection{PERFORMANCE AND ACCURACY RESULTS}
\label{secPerformance}
\begin{figure}[tb] 
\includegraphics[width=\columnwidth]{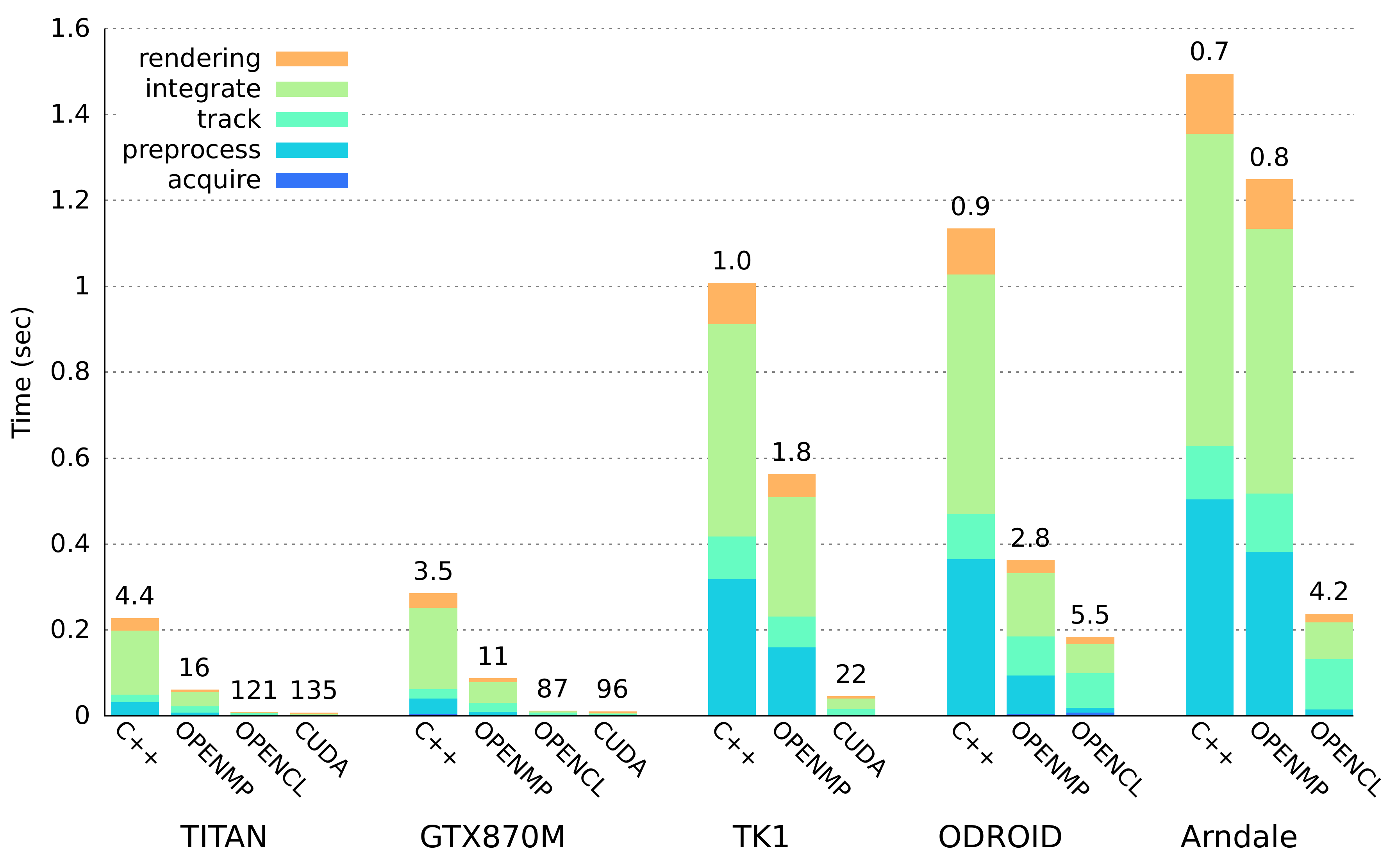} 
\caption{High-level SLAM building block elapsed times. FPS is reported on top of each histogram.} 
\label{figPlatform} 
\end{figure}

Default SLAMBench parameters are used for the experiments as in \cite{KFusion}: 
integration rate is set at 2 frames, the voxel volume is $256$x$256$x$256$, 
the volume size is $4.8$x$4.8$x$4.8$ $m^3$, 
rendering rate is set at 4 frames and the input image is halved during the preprocessing, i.e. $320$x$240$.
All the experiments are run on the living room trajectory 2 of the ICL-NUIM dataset. 
\begin{table*}[htb]
\begin{center}
\begin{tabu}{cccc|cccc|ccc|ccc|ccc}
\hline

\rowfont{\bfseries\footnotesize}  \multicolumn{4}{c}{TITAN} &  \multicolumn{4}{c}{GTX870M} &  \multicolumn{3}{c}{TK1} & \multicolumn{3}{c}{ODROID} & \multicolumn{3}{c}{Arndale}\\
\rowfont{\bfseries\footnotesize}  C++ & OMP & OCL & CUDA & C++ & OMP & OCL & CUDA & C++ & OMP & CUDA & C++ & OMP & OCL & C++ & OMP & OCL\\
\hline 
\rowfont{\footnotesize}  2.07 & 2.07 & 2.07 & 2.07 & 2.07 & 2.07 & 2.07 & 2.07 & 2.06 & 2.06 & 2.07 & 2.06 & 2.06 & 2.01 & 2.06 & 2.06 & 2.07\\

\hline

\end{tabu}
\end{center}
\caption{Absolute trajectory error (ATE) in centimeters.}
\label{tabATE} 
\end{table*}
Table~\ref{tabATE} demonstrates a consistent maximum ATE of between 2.01cm  on the ODROID OpenCL implementation, and 2.07 cm on other platforms/implementations was computed over all runs. 
These errors are comparable with  \cite{2014Handa}.
The \textit{reduce} kernel gives rise to small numerical differences as it
performs non-associative summations leading to different rounding errors. 
Fig.~\ref{figPlatform} shows the performance achieved across different platforms for all available languages. 
The FPS achieved are reported on top of each histogram.  
The y axis represents the time in seconds to process one frame. 
The TITAN desktop achieves the highest frame rate running at 135 FPS. 
Using their on-die GPU, the embedded TK1 platform almost reaches real-time with ca. 22 FPS, 
while the ODROID and Arndale boards achieve an average frame rate of 5.46 and 4.21 respectively.
Using the CPU cores only, via OpenMP, the 8-core ODROID platform 
achieves 2.76 FPS while the Arndale and TK1 reach 0.8 and 1.78 respectively. 

Despite our efforts to make the OpenCL and CUDA versions as similar as possible, 
we observe a 10\% difference of performance in favour of CUDA. 
An unexpected result is the difference in the proportion of time spent in each kernel between OpenCL and 
CUDA in the desktop and laptop histograms, see next paragraph.
We are currently investigating why such a wide difference exists and why \textit{renderDepth} and \textit{renderTrack} scale so badly in CUDA. 

\begin{figure}[tb] 
\includegraphics[width=\columnwidth]{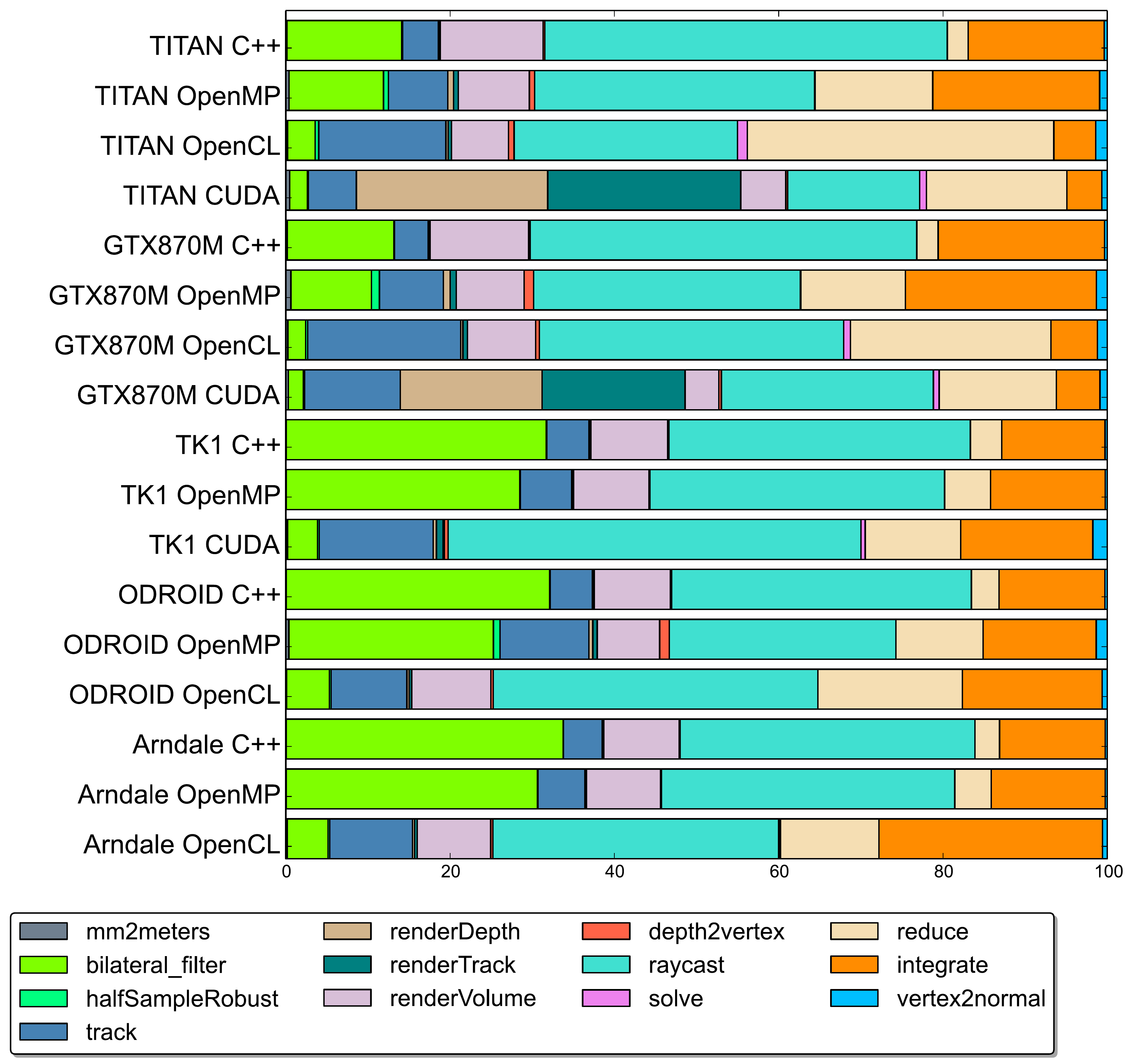} 
\caption{Percentage of time spent in each kernel.} 
\label{figKernel} 
\end{figure}

Fig.~\ref{figKernel} shows the percentage of time spent in each kernel and 
how this changes over different platforms/implementations. 
Ideally, an application containing kernels that scale perfectly will have 
the same proportions on all the histograms of a given platform. 
If a kernel scales well, i.e. better than other kernels in the average, 
then it would reduce its proportion on a histogram that contains more computational units for a given platform. 
As an example, if we consider the 8-core ODROID and 
we compare the C++ and OpenMP histograms we can see that the \textit{bilateralFilter} and 
the \textit{raycast} scales well.  
In contrast the \textit{reduce} and the \textit{track} kernels scale badly, whereas the \textit{integrate} is average. 
We can see this effect on all the platforms. 
The proportions stay roughly unchanged on the 2-core Arndale  because of the limited number of cores. 

There is a correlation between Table~\ref{tabKernels} and the qualitative scaling factors observed in Fig.~\ref{figKernel}. 
The reasons why \textit{bilateralFilter} scales well are threefold: 
first it operates a stencil pattern; second it loads/stores 2D images which implies that data can fit into caches 
exploiting data locality; 
third the percentage of the program taken by this kernel is not
negligible which means that this kernel has room for speedup.
The \textit{reduce} kernel does not scale linearly because it follows a reduce pattern; 
despite tree-based reduction, this kind of pattern suffers 
on highly parallel devices as can be observed in the CUDA and OpenCL histograms.
Since \textit{integrate} manipulates 3D data it takes a performance hit on platforms with reduced data cache size, i.e. embedded devices, 
while scaling nicely on TITAN and GTX780M.
The \textit{raycast} bar is approximately constant across the histograms. 
This kernel also manipulates 3D data and, as a consequence, has high
memory bandwidth demand and has intensive computation due to the stencil pattern it operates. 
The high number of load/store operations is  balanced by the high computation needed. 

We can also notice that small kernels, which are  not visible in the C++ histograms, 
if they scale badly, will appear in histograms where massive parallel resources are deployed. 
An example is the CPU-only \textit{solve} kernel. 

\subsection{ENERGY CONSUMPTION ANALYSIS: ODROID-XU3 CASE STUDY}
\label{secPower}
Anecdotally it is sometimes suggested that GPU based applications are power-hungry, 
indeed in many systems they do draw a significant amount of the system power, however the GPU may also carry a disproportionate amount of the workload.  
We use the ODROID device introduced in Sec.~\ref{secDevices}
to evaluate the energy efficiency of  its various processing elements in the context of a SLAM application. The 
platform  has on-board voltage/current sensors and  split power rails, which allows the power consumption of the "big" Cortex-A15 cores,
 the "LITTLE" Cortex-A7 cores, GPU and DRAM to be measured individually.  The sensors can be read by the SLAMBench application 
to provide measurements at various points of execution which can be logged or plotted interactively  in the GUI. 
Energy consumption is only explored on the ODROID because consistent power analysis tools are not present on all devices. 

We use SLAMBench along with the platform's energy monitors to measure energy/power consumption and  
performance using the C++, OpenMP and OpenCL implementations; 
for C++ tests using both Cortex-A15 and Cortex-A7, for OpenMP using all eight cores or just Cortex-A7 and 
OpenCL using only the larger GPU device and either a Cortex-A15 or Cortex-A7. 
These tests are conducted using Process-every-frame mode, and consequently show the relative energy efficiency 
of the various implementation/resource combinations for the same amount of computation.  
Currently, the ODROID's Linux kernel only supports a performance driver 
which prefers to shift threads onto the Cortex-A15 cores, to deliver high performance. 
In order to prevent this, and to force the use of Cortex-A7 cores, we set the Cortex-A15 cores to be offline where appropriate. 

\begin{figure}[tb] 
\includegraphics[width=\columnwidth]{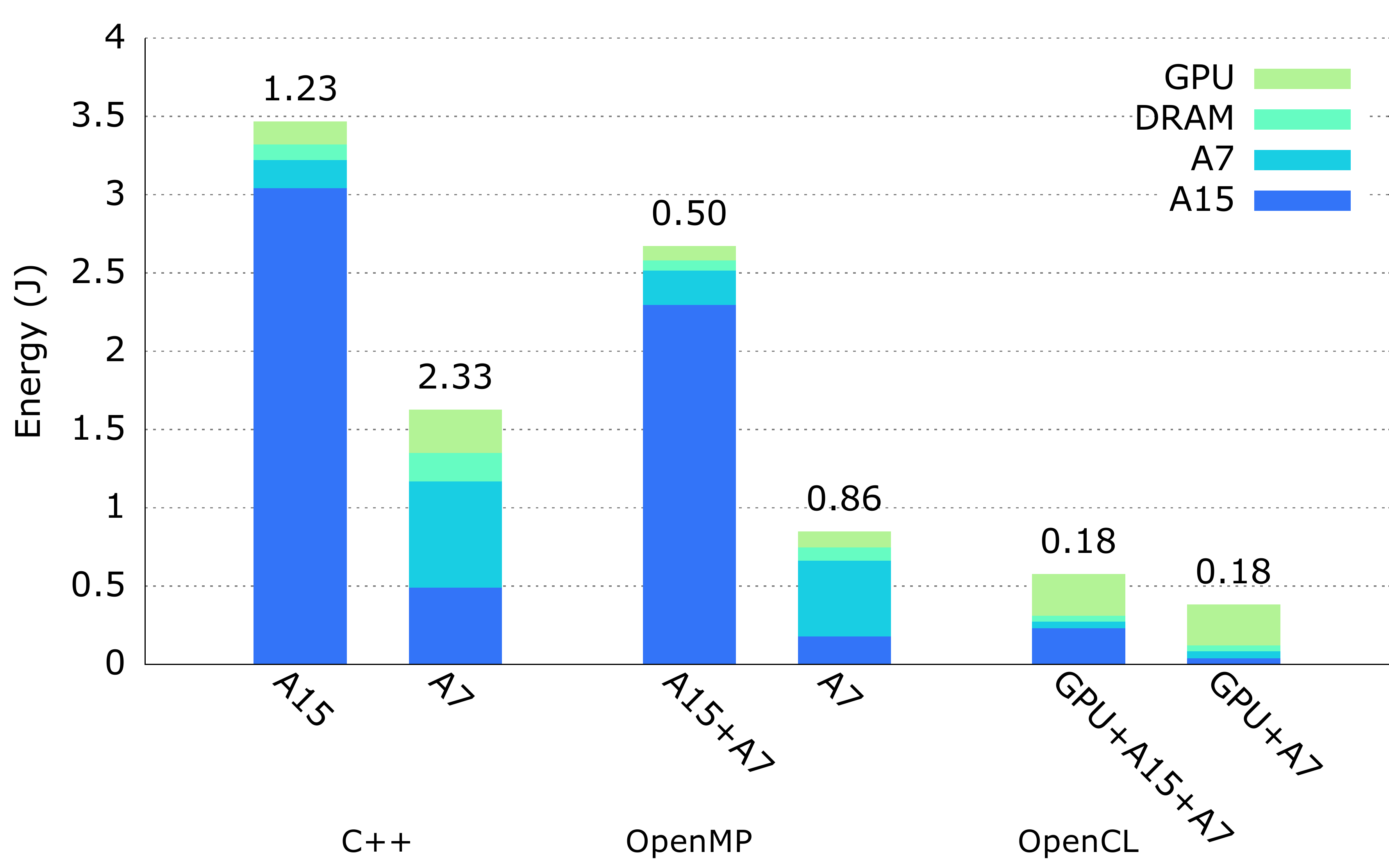} 
\caption{Energy usage per frame utilising various hardware resources on the ODROID, mean time per frame marked}
\label{odroid_energy} 
\end{figure} 

Fig.~\ref{odroid_energy} clearly demonstrates that the OpenCL/GPU implementation uses significantly less energy  
than approaches that rely solely on the conventional CPU cores. 
Furthermore we can see that, for  OpenCL, the performance benefit from using the Cortex-A15 over the Cortex-A7 is negligible, but
the energy consumed is 50\% higher (507.28 compared with 337 Joules), there is consequently no benefit in this application of using the high performance CPU.
 Implementation using OpenMP on Cortex-A7 also gives an improvement in energy over a single 
 core C++ implementation, assuming a race to sleep strategy is employed whereby data is processed as fast we can then we switch a low power mode, 
 indicating potential energy savings are achievable in SLAM applications by using parallel implementations.
There is a significant energy usage for the Cortex-A15 even when it has been configured as offline, 
this would point to the operating system not forcing the cores to be powered down and indicating a greater reduction in energy consumption may be achievable when they are not in use. 
  
A SLAM system will fail if the inter-frame movement of the camera is too great, as it will fail to track the camera pose. 
We can see for the ODROID we would typically drop a significant number of frames using our default settings, 
of voxel size, image size, integration rate and rendering rate. 
SLAMBench allows the user to run at the rate of the system, 
dropping frames where appropriate. 
The user can then explore the design space to achieve an optimal ATE. 
For example on an embedded system they may choose not to create a coarser model, render our model as it is  created - this is typical for a robotics application -   
or update the model less often, all of which will, typically, deliver substantial reduction in dropped frames and, within limits, the ATE.

Parallel approaches offer a significant energy, and performance, improvement over single threaded approaches. The Mali GPU provides a significantly improved energy efficiency when compared with conventional cores for this SLAM application, 
but care must be taken to ensure that conventional CPU threads are scheduled on the most appropriate conventional cores, 
and that unused cores are put into an appropriate low power state.

\section{CONCLUSIONS AND FUTURE WORK} 
\label{CONCLUSIONS} 
SLAMBench is the first SLAM benchmark which is concerned
not just with accuracy but also computational performance and  energy consumption.  
We show how SLAMBench allows a user to perform a fair comparison of hardware accelerators,  
software tools and novel algorithms in a modern dense SLAM framework, where 
our framework for automatically comparing estimated trajectories
against ground truth ensures that estimation accuracy is maintained
and separated from performance comparison.
Each SLAMBench kernel is characterised by its parallel pattern and 
the weight it has in the SLAM computation;
swapping module implementations is relatively straightforward so
that certain parts of an algorithm can be isolated and improved.

We present performance results on state-of-the-art desktop, laptop and embedded devices. 
Four configurations achieved super real-time FPS in the desktop and 
laptop space with a peak performance of 135 FPS.  
The Tegra K1 achieves 22 FPS, which is impressive for an embedded device, though
this board is not optimised for energy usage, 
which may restrict its applicability in robotics. 
In contrast, the ODROID-XU3 achieves 
5.5 FPS for 
2.1 
Watts average dissipation. 
All of these  performance figures leverage the efficiency of GPUs, showing 
how effective heterogeneous computing can be for SLAM on a wide range of modern devices. 

We have also presented the first SLAM energy consumption evaluation 
on the ODROID-XU3 embedded platform which is suitable for robotics,
analysing the performance/power trade-off at a fixed accuracy.
The Cortex-A7/MALI provides the same FPS as the Cortex-A15/MALI with a 50\% lower  
energy envelope.
This research paves the way for systematic power dissipation evaluation using SLAMBench. 
Future work will include the addition of performance counter based energy monitoring, 
on devices that provide hardware counter APIs, by inclusion of PAPI within the framework as outlined in \cite{2012Weaver}
 
SLAMBench facilitates design-space exploration. 
A number of algorithm parameters, such as the voxel density or the TSDF distance $\mu$, 
are available in the SLAMBench user interface. 
Exploring how these parameters improve the SLAM application 
while simultaneously changing compiler optimisation parameters, 
such as the OpenCL work-group size or thread-coarsening \cite{2013Magni}, 
may have a high impact on performance/power/accuracy metrics. 
 
Finally, SLAMBench provides  
the necessary foundation for investigating domain-specific optimisations.  
We plan to build our own DSL,   
targeting high performance, low-power SLAM applications, and believe  
that it will inspire similar efforts from relevant communities.  
SLAMBench is intended to develop over time and it is expected that it will include a larger set of 
SLAM kernels in later versions. 
A key step would 
be to integrate kernels that allow KFusion to be 
more scalable in terms of the size of the scene to be reconstructed,  
for example, by adding alternative and complementary data structures such as point-based fusion \cite{2013Keller},  
octrees \cite{2013Zeng}, voxel hashing-based representation 
\cite{2013Niessner} and moving volumes, as used in Kintinous \cite{2012Whelan}.

\addtolength{\textheight}{-12cm}   
 
\section*{ACKNOWLEDGMENTS} 
We acknowledge funding by the EPSRC grant PAMELA EP/K008730/1.  
M. Luj\'an is funded  by a Royal Society University Research Fellowship.
We thank G. Reitmayr for the original KFusion implementation; 
G. S. Shenoy, A. Handa, B. Franke, D. Ham, 
and PAMELA Steering Group for discussions.
 
\bibliography{pamela} 
\bibliographystyle{abbrv} 
 
\end{document}